\documentclass{article}

\usepackage{arxiv}

\usepackage[utf8]{inputenc} 
\usepackage[T1]{fontenc}    
\usepackage{hyperref}       
\usepackage{url}            
\usepackage{booktabs}       
\usepackage{amsfonts}       
\usepackage{nicefrac}       
\usepackage{microtype}      
\usepackage{lipsum}
\usepackage{graphicx}
\usepackage{amsmath}

\graphicspath{ {images/} }

\title{Dirichlet uncertainty wrappers for actionable algorithm accuracy accountability and auditability}

\author{
  Jos\'e Mena \\
  Eurecat, Centre Tecnol\`ogic de Catalunya\\
  Universitat de Barcelona \\
  \texttt{jose.mena@eurecat.org} \\
   \And
 Oriol Pujol \\
  Dept. Matem\`atiques i Inform\`atica\\
  Universitat de Barcelona\\
  \texttt{oriol\_pujol@ub.edu} \\
   \AND
 Jordi Vitri\`a \\
  Dept. Matem\`atiques i Inform\`atica\\
  Universitat de Barcelona\\
  \texttt{jordi.vitria@ub.edu} \\
}

\begin{document}
\maketitle

\begin{abstract}
Nowadays, the use of machine learning models is becoming a utility in many applications. Companies deliver pre-trained models encapsulated as application programming interfaces (APIs) that developers combine with third party components and their own models and data to create complex data products to solve specific problems. The complexity of such products and the lack of control and knowledge of the internals of each component used cause unavoidable effects, such as lack of transparency, difficulty in auditability, and emergence of potential uncontrolled risks. They are effectively black-boxes. Accountability of such solutions is a challenge for the auditors and the machine learning community. In this work, we propose a wrapper that given a black-box model enriches its output prediction with a measure of uncertainty. 
  By using this wrapper, we make the black-box auditable for the accuracy risk (risk derived from low quality or uncertain decisions) and at the same time we provide an actionable mechanism to mitigate that risk in the form of decision rejection; we can choose not to issue a prediction when the risk or uncertainty in that decision is significant. Based on the resulting uncertainty measure, we advocate for a rejection system that selects the more confident predictions, discarding those more uncertain, leading to an improvement in the trustability of the resulting system. We showcase the proposed technique and methodology in a practical scenario where a simulated sentiment analysis API based on natural language processing is applied to different domains. Results demonstrate the effectiveness of the uncertainty computed by the wrapper and its high correlation to bad quality predictions and misclassifications.
\end{abstract}


\section{Introduction}

Algorithmic decisions coming from machine learning models are being used everyday in all sectors. With this broad adoption of technology there exists legitimate concerns on the use of these models when they effectively impact lives, such as on healthcare \cite{product:DigitalReasoning}, education  \cite{product:Turnitin}, or employment \cite{product:AvrioAI} \cite{product:Ideal}. The increment on reports on failures and unintended effects \cite{Amodei:2016concrete} \cite{bostrom:2009ethics} \cite{sculley:2014nips} is mobilizing society to ask for algorithm accountability and transparency \cite{angwin:2016accountable} \cite{goodman:2016accountable} and the scientific community to give an answer to these challenges. 

Algorithm accountability and transparency aim at promoting trust on machine learning models by disclosing their intended use, identifying the potential impact and legal risk of the application of those solutions, and mitigating their unintended consequences. In 2016, the community for Fairness, Accountability, and Transparency in Machine Learning (FATML) released a set of five principles and associated suggested social impact statements and guidelines \cite{fatml:2016principles} to "help developers and product managers design and implement algorithmic systems in publicly accountable ways". These five principles accounted for the following dimensions: responsibility, explainability, accuracy, auditability, and fairness. Relevant to this article are the dimensions of accuracy and auditability. 

Accuracy concerns the direct impact of errors in the solution. In this dimension, the effect of unreliable and low-quality predictions are to be understood and eventually mitigated when it resolves in potential harm or risk. In this context, uncertainty in the decisions naturally emerges as a relevant element. Uncertainty in machine learning has always been a concern of Bayesian models in machine learning \cite{koller2009probabilistic}. In recent years with the resurgence of deep learning, the reinterpretation of some existing mechanisms such as dropout or stochastic mechanisms as Monte Carlo approaches has broadened the use of these techniques for accounting for uncertainty in deep models \cite{gal2016uncertainty}. Uncertainty can be categorized into epistemic and aleatoric uncertainty. Epistemic uncertainty accounts for the uncertainty that appears in the model parameter inference from the data, i.e. the variability in the model parameters for a particular finite dataset. This uncertainty can be reduced as more data is considered. In real production scenarios, this kind of uncertainty is not available since the deployment stage usually encompasses the selection of a single solution, i.e. the model that achieves the maximum a posteriori probability given a dataset. The second kind of uncertainty is called aleatoric uncertainty and corresponds to the uncertainty inherent in the data. For example, this may account for ambiguity in semantics in natural language processing systems. This type of uncertainty can not be reduced even with more data.

Although these are relevant sources for reporting the reliability of a decision, we need a way to make this uncertainty actionable. In this respect and following the principle of risk aversion in which "it is better not to answer than produce an answer that might harm with high probability" we advocate for using rejection techniques (aka selective prediction). In this sense, uncertainty is a clear candidate for a rejection measure: we might opt for abstaining on producing an answer when predictions are very uncertain. In literature, we find examples of different rejection functions \cite{DBLP:conf/icml/GeifmanE19} \cite{DeStefano:2000:RRQ:2220396.2220558}, even using uncertainty \cite{NIPS2017_7073} as the rejector. The main limitation of these methods in the context of this article is that they require to train the rejector together with the classifier and need access to the internals of the model. 

Although actionable uncertainty is clearly a path to follow for algorithmic accuracy accountability, this becomes a problem when one is not in control of all steps and/or components in a data product. It is often the case that a data-based solution is composed of third-party closed products or Software-as-a-Service (SaaS) prediction APIs mixed with one's algorithms. This disparity makes the risk control of the solution complex since one has no access to the internals of some of the components nor even the data they have been trained with. This real common scenario motivates the importance of {\it auditability} as another relevant dimension for data product accountability. In general terms, auditability refers to the capability of an algorithmic solution to be probed, monitored, and understood. Most of the solutions proposed for mitigating adverse outcomes in accountability require access to the model internals \cite{menon:2018fairness} \cite{rudin:2018white}. However, in the context of explainability, several proposals operate directly by considering the model a black-box, for example in LIME \cite{DBLP:journals/corr/RibeiroSG16}. This is an example of a {\it wrapper} system, that takes a black-box model and endows it with a new feature. Wrappers are essential tools for enabling auditability of a system concerning different accountable dimensions. Wrappers have a long tradition in machine learning, especially in the field of feature selection \cite{yan2015feature} \cite{pudil1994floating}. In some sense, these works can be regarded as the first steps on explainability.

In this article, we propose (i) a deep learning wrapper based on the Dirichlet distribution that allows embedding any black-box classification system, as long as a distribution of classes is provided, with uncertainty features and (ii) we exploit the measured uncertainty advocating for rejection mechanisms to enforce accountable accuracy. We showcase the proposal in sentiment analysis on a natural language processing scenario where we consider sentiment predictions are given by an immutable undisclosed API used in a different domain than the one the API has been trained. This reflects the real case in which a third-party API is used in a data product where we do not know the model internals nor the training data distribution. However, despite this unknown information, we still require responsible algorithm accountability in our solution.

In section 2, we introduce the method proposed for building an uncertainty wrapper around a black-box model. In section 3, we describe how to obtain an uncertainty score from the wrapper output. Section 4 introduces the concept of rejection and rejection performance metrics. In section 5, we showcase the proposed method in four different scenarios for sentiment analysis in natural language processing. The results obtained corroborate the importance of the rejection method and show the success of the proposed methodology. Finally, section 6 concludes the article. 


\section{Building an uncertainty wrapper}
The present section describes a deep learning technique for building a wrapper that endows any black-box model with uncertainty capabilities. We show how this uncertainty can assess the fitness of the black-box model to the target application domain and can be used for rejecting the more uncertain points. Hence, by removing the more uncertain points, the method increases the accuracy of the remaining predictions for the new domain and complies with the accuracy risk mitigation principle. 

\subsection{Problem description}

The starting point of the present work's use case is a black-box model pre-trained to solve a given NLP classification task. As an input, the model receives a text, composed by $N$ words ${w_1,w_2,...,w_N}$, applies the learned function $f$ to the text and outputs a probability distribution among the labels, ${y_1,...,y_C}$, where $C$ is the number of possible classes, indicating on each position the probability of the text of belonging to that given class.

Nowadays, there is plenty of these black-box models. For example, online APIs offer prediction models for many different tasks, including NLP. APIs like the one offered by Google in their Cloud Natural Language API \footnote{\url{https://cloud.google.com/natural-language/}}, or Turbo NLP \footnote{\url{https://turbonlp.com/}}, among others, offer methods for sentiment analysis, text summarization, classification, and so forth. In this scenario, there are two things to remark. On the one hand, we do not have access to the internals of the model. In practical terms, we ignore the type of model implemented, the learned parameters, or even if there is a unique predictor, or there are different models combined to fulfil the task. On the other hand, in most of the cases, we neither know the data used to train the model. We only have an input text and an output prediction.

In this setting, imagine that an analyst wants to use these interfaces for building a product, let us say for sentiment analysis, applying them to her problem and domain. Alternatively, think about a company that has developed one of these models for a customer and wants to know if they can sell it to other customers who will apply it to different domains. To what extent can they trust the output of the model in their new problem and domain? Maybe the target dataset is very different from the original. Maybe there are words that the model has never seen during the training. Or maybe the meaning of those words change from one domain to another. By directly using the output of the classifier, we may incur in an unwanted and unintended risk. Indeed, a way for measuring and mitigating this risk is needed.

\subsection{Modelling the distribution}

The type of uncertainty used in this paper is the so-called heteroscedastic aleatoric. Differently from other types of uncertainty described in the literature, \cite{kendall2017uncertainties}, this type of uncertainty measures the variability in the output predictions caused by the noise inherent in the input data. In this case, we can not work with the epistemic uncertainty as it is related to the parameters of the model and recall that here we do not have access to them.

In a classification setting, the output is a probability distribution $p(y|X, w^{*})$ over the different classes to predict, $y$, given an input text $X$ and the pre-trained model with parameters $w^{*}$. We model this output by a random variable to measure the variability that the data noise causes in the output. With this approximation, we propose different metrics derived from the output of the random variable and use them as a proxy for the associated uncertainty. In our modelling, we assume that the output probability distribution comes from a Dirichlet probability density function.

In previous works, \cite{kendall2017uncertainties}, they use independent Gaussian random variables to model the pre-activation value of the logits\footnote{The values before the softmax layer} in a deep learning architecture. This is done in order to capture the noise in each class. However, this approach does not conform to the constraints in our setting. Having access to the logits before the softmax breaks the black-box assumption and requires access to the internals of the model. Besides, the model exclusively works for deep learning settings where the last layer is a softmax.

In order to comply with the pure wrapper approach, there are several constraints to observe: First, we need to exclusively operate on the output of the classifier. We are not allowed to use any intermediate or internal value of the black-box model as we need to be agnostic to that model. Next, since we are modelling a distribution over the output classes, the probability constraints apply (each value in the output belongs to the interval $[0,1]$, and the sum adds up to one). The Gaussian distribution is not suitable for that effect. A more natural approach is to consider the output distribution coming from a {\it Dirichlet} probability density function. This is, the output distribution is modelled 

\begin{equation}
p(y|X, w^{*}) \sim Dir(\boldsymbol{\alpha})
\end{equation}

where $\boldsymbol{\alpha}$ is the concentration parameters that control the Dirichlet distribution, $\boldsymbol{\alpha} = \left \{ \alpha_{1},...,\alpha_{C} \right \}$. In the literature, we may find works that propose using a Dirichlet distribution like in \cite{gast2018lightweight, NIPS2018_7936, chen2018variational}, where they propose different alterations of the original model to learn the output distribution. The problem with those works is that they do not observe the black-box constraint here imposed on not having access to the original model.

\subsection{Decomposability of the Dirichlet concentration parameter}

At this point, a question arises: given the output of the black-box, how can we relate the output of the classifier with the concentration parameter in the Dirichlet distribution? 

We propose a decomposition of the concentration parameter in two terms. To that effect, we recall some basic statistics of the Dirichlet distribution. Given a Dirichlet random variable $\boldsymbol{x}\in \mathbb{R}^C$ with concentration parameter $\boldsymbol{\alpha}\in \mathbb{R}^C$, the expected value of the distribution is defined as
$$\mathbb{E}(\boldsymbol{x}) = \frac{\boldsymbol{\alpha}}{\sum\limits_{i=1}^C \alpha_i}.$$ 

Observe that the expected value has the same properties as a probability distribution and that the output of the black-box $\boldsymbol{y}\in \mathbb{R}^C$ is already a probability distribution. In this sense, we could directly use the output as the concentration parameter. However, each term of the concentration parameter is not necessarily constrained to the interval $[0,1]$. Let us introduce a new scalar parameter, $\beta \in \mathbb{R}$ that will model this difference, such that

$$\boldsymbol{\alpha} = \beta \boldsymbol{y}.$$

Observe that $\beta$ is a scalar value that varies with each input $\boldsymbol{x}$ and ensures that for any value of $\beta$ the expected value remains constant, and thus guarantees that the expected output displays the same behaviour as the original black-box classifier. 

\begin{figure}[!ht]
\centering
\caption{Dirichlet distribution in 3 dimensions for different $\beta$ values given a prediction of [0.25,0.25,0.50].}
\minipage{0.3\linewidth}
  \centering
  \includegraphics[width=\linewidth]{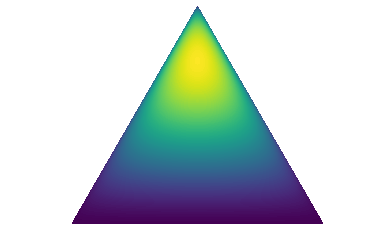}
  \caption{$\beta = 5$}\label{fig:dirichlet_beta5}
\endminipage\hfill
\minipage{0.3\linewidth}
  \centering
  \includegraphics[width=\linewidth]{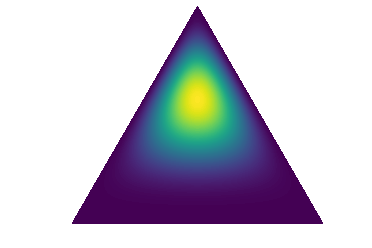}
  \caption{$\beta = 10$}\label{fig:dirichlet_beta10}
\endminipage\hfill
\minipage{0.3\linewidth}
  \centering
  \includegraphics[width=\linewidth]{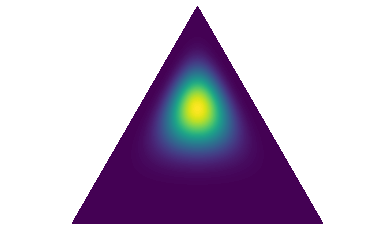}
  \caption{$\beta = 20$}\label{fig:dirichlet_beta20}
\endminipage\hfill
\end{figure}

This decomposition has the property that each component is easily interpretable. While the output of the black-box classifier stands for the mean, parameter $\beta$ accounts for the spread of the distribution. An example of the effect of varying this parameter in a three dimensional Dirichlet distribution is shown in Figures \ref{fig:dirichlet_beta5} to \ref{fig:dirichlet_beta20}. Observe that the higher the value of $\beta$, the more pointy the distribution is.

This decoupling allows to effectively isolate the contribution of the black-box and the contribution that remains to be computed, i.e. the value of parameter $\beta$. We propose to use this decomposition to create a wrapper, that takes as an input the value of $\boldsymbol{y}$ from the original black-box classifier and learns the value of $\beta$ given the data at hand. The proposal uses a deep learning regressor that takes as input the same data that enters the black-box and outputs the value of $\beta$. 

\begin{figure}[ht!]
\centering
\caption{Model used to estimate the aleatoric uncertainty from the original black-box model}
\includegraphics[width=\linewidth]{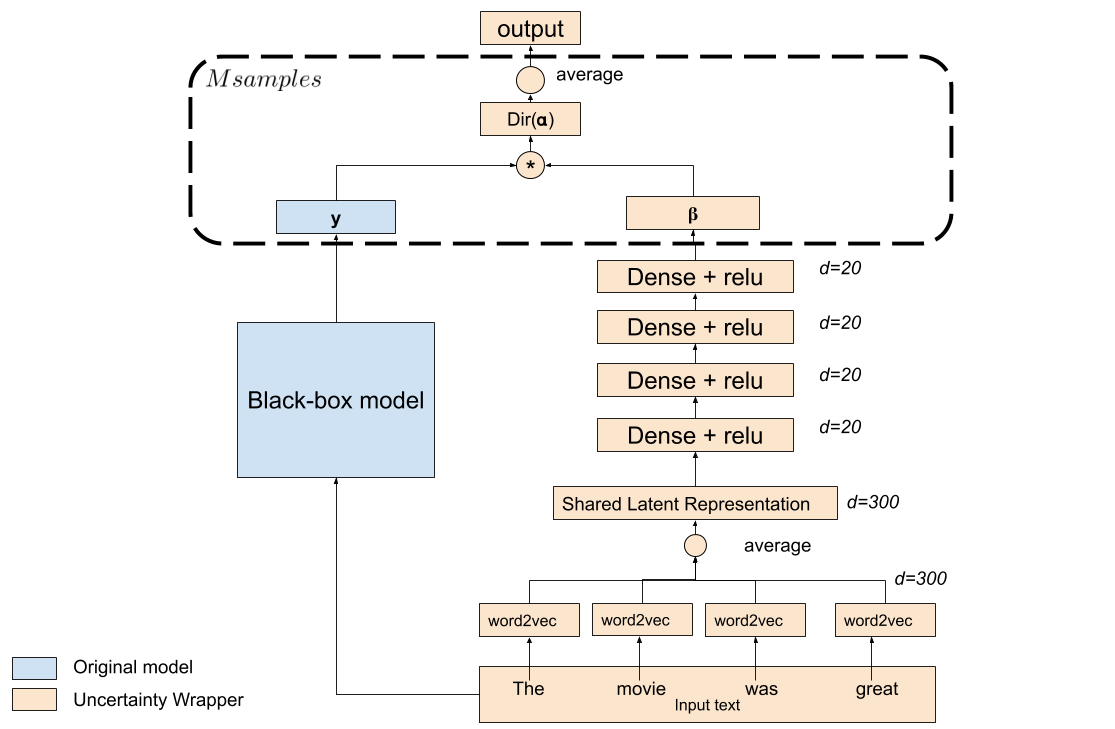}
\label{fig:model}
\centering
\end{figure}

Figure \ref{fig:model} shows the integration of the wrapper (in light orange color) with the black-box classifier (in light blue color). The architecture used in this figure corresponds to the one that will be used in the experimental section for the case of sentiment analysis for natural language processing. In this particular case, as an example, a sentence is represented as the average value of the embedding of each sentence word using word2vec embeddings\footnote{Other embeddings can be used to this effect, such as ELMO, Bert, among others.}. This representation in feed to four dense layers of twenty neurons, each one with ReLu activation functions. Finally, a dense layer compresses the outputs to a single regression value, that will be used as parameter $\beta$. Note that the proposal is general and the same principles can be used in any other domain besides NLP as long as the wrapper includes a representation of the input expressive enough to be able to capture this uncertainty.

\subsection{Inference in the Dirichlet setting}

The expected value of the classification probabilities can be approximated using Monte Carlo sampling \cite{kendall2017uncertainties} from the learned Dirichlet distribution for each sample, $\hat{y}_{m,i} \sim Dir(\alpha_i)$ as

\begin{equation}
\mathbb{E}[\hat{y}_i] = \frac{1}{M}\sum^{M}_{m=1} \hat{y}_{m,i}
\label{eq:expected}
\end{equation}

Where M is the number of samples drawn. This is used to define the loss function for our learning stage. Given a set of $N$ training samples, we propose to use a regularized version of the cross-entropy loss function as follows,

\begin{align}
    \mathcal{L}(W) = - \frac{1}{N}\sum^{N}_{i=1}\frac{1}{C}\sum^{C}_{c=1}y_{i,c}\log{\mathbb{E}[\hat{y}_i]_c} + \lambda\|\beta\|_2
    \nonumber\\
    = -\frac{1}{N}\frac{1}{C}\sum^{N}_{i=1}\sum^{C}_{c=1}y_{i,c}\log\big({\frac{1}{M}\sum^{M}_{m=1}\hat{y}_{m,i,c}}\big) + \lambda\|\beta\|_2.
\end{align}

Observe that we introduce the norm of the $\beta$ value in the minimization function. This term is required since the unregularized cross-entropy forces the value of $\beta$ to grow unbounded. By adding this term, we can control its growth. Governing the trade-off, we introduce parameter $\lambda$. The value of parameter $\lambda$ is not critical besides extreme cases, and yields very similar results for reasonable values of the stable interval.

Once the network is trained, we effectively obtain a wrapper that endows the black-box output distribution with a Dirichlet probability density distribution on top that enables defining the uncertainty score.

\section{Obtaining an uncertainty score from the wrapper} \label{sec:unc_score}
After training the model, the uncertainty wrapper provides a method that given an input text and its original prediction, combines the original prediction $\boldsymbol{y}$ with the uncertainty parameter $\beta$ to obtain the concentration parameter $\boldsymbol{\alpha}$ for the Dirichlet distribution. Next, we can sample from this distribution to obtain different probability distributions of the classifier output, $\hat{y}_m \sim Dir(\boldsymbol{\alpha})$, and study their variability.

Although being able to sample from the learned distribution is an interesting feature of the proposed wrapper, its final goal is to obtain a numerical score for the uncertainty. In the literature \cite{gal2016uncertainty} we find different approaches for obtaining a numerical value out of the output probability distributions in classification. In the present article, we explore two of them, namely {\it variation ratios} and {\it predictive entropy}. Both use Monte Carlo simulation sampling from the obtained Dirichlet function.

\textit{Variation ratios} measures the variability of the predictions obtained from the sampling \cite{freeman1965elementary}. This heuristic is a measure of the dispersion of the predictions around its mode:
\begin{equation}
c^* = \arg\max_{c=1,...,C} \sum_{t} 1{[y^{t} = c]}
\end{equation}
using the M number of times it is sampled,
$f_x = \sum_{t} 1{[y^{t} = c]}$, and the variation ratio for an input $x$ is $VR = 1 - \frac{f_{x}}{M}$

Alternatively, \textit{predictive entropy} is based on information theory, and it considers the average amount of information contained in the predictive distribution. Those results with low entropy values correspond to confident predictions, whereas high entropy leads to large uncertainty.

Note that the output of the black-box model $\boldsymbol{y}$ already describes a probability distribution, so one could compute its predictive entropy and obtain a measure of its uncertainty as follows,

\begin{equation}
\mathbb{H} = -\sum_{c} y_c\log{y_c}
\end{equation}

However, the wrapper allows to model the variability of the black-box output distribution, and thus we can compute a predictive entropy that not only considers the entropy of the output but also the confidence on that entropy. Again, given $\hat{y}_m \sim Dir(\beta \boldsymbol{y})$ and its expected value as defined in Equation \ref{eq:expected}, the sampled predictive entropy is defined as

\begin{equation}
\mathbb{H} = -\sum_{c} \mathbb{E}[\hat{y}]_c\log{\mathbb{E}[\hat{y}]_c}.
\end{equation}

In this work, we demonstrate how the proposed method captures better the uncertainty compared to the predictive entropy of the original model.

\section{Using uncertainty for rejection}

So far, the proposed method delivers an uncertainty score given an input and the associated prediction from a pre-trained black-box classifier. In this section, we describe a way of making uncertainty actionable by recalling the concept of rejection or selective prediction. Rejection is a mechanism such that given a particular metric, usually related to the confidence in the decision, we opt not to output a prediction if the related metric value is below some threshold. 

In our proposal, we use the computed uncertainty as this rejection metric, such that those predictions with significant uncertainty are not given as an output. In the context of our use case, the initial hypothesis is that texts with high uncertainty are prone to be misclassified by the black-box model.

In order to use the uncertainty score for evaluating the performance of the black-box in a new dataset, we first proceed to obtain the predictions applying the original model. Then for each pair of text and prediction, we obtain the associated uncertainty score using the wrapper. Next, we sort the predictions based on the uncertainty score, from more to less uncertain. From that ordering, we set the rejection threshold that marks where to start trusting the classification model.

\begin{figure}[ht!]
\centering
\caption{Rejection performance metrics as proposed in \cite{condessaKB15}}
\includegraphics[width=0.8\linewidth]{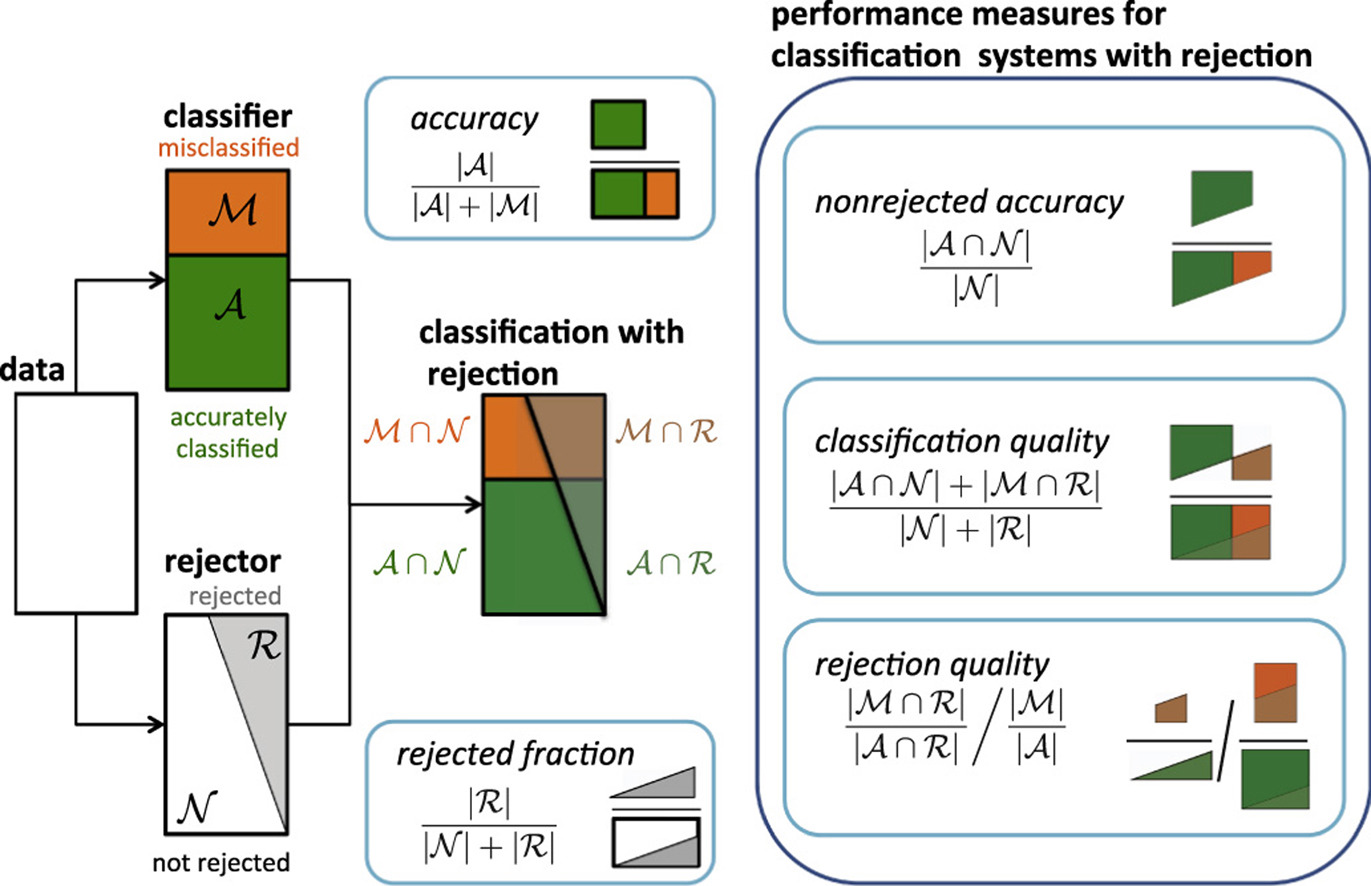}
\label{fig:condessa_metrics}
\centering
\end{figure}

To evaluate the rejection metric, we split the dataset using two criteria: whether the method \textbf{R}ejects the data point or \textbf{N}ot; and whether the point is \textbf{A}ccurately classified\footnote{We measure accuracy by comparing the test label with the category that holds a higher probability}, or \textbf{M}isclassified named as R, N, A or M respectively. Using this terminology, we follow the guidelines in \cite{condessaKB15} for rejection quality metrics. We have three quality metrics, illustrated in \ref{fig:condessa_metrics}:

\begin{itemize}
    \item {\bf Non-rejected Accuracy} measures the ability of the classifier to classify non-rejected samples accurately. It is computed as follows,
    
    $$NRA = \frac{\left | A \bigcap N \right |}{\left | N \right |}$$
    
    \item {\bf Classification Quality} measures the ability of the classifier with rejection to classify non-rejected samples accurately and to reject misclassified samples. It is computed as follows,
    
    $$ CQ = \frac{\left | A \bigcap N \right | + \left | M \bigcap R \right | }{\left | N \right | + \left | R \right |}$$
    
    \item {\bf Rejection Quality} measures the ability to concentrate all misclassified samples onto the set of rejected samples. It is computed as follows,
    
    $$ RQ = \frac{\left | M \bigcap R \right | \left | A \right |}{\left | A \bigcap R \right | \left | M  \right | }$$
\end{itemize}

A good rejection point will show a trade-off between the three metrics, being able to divide misclassified predictions from right ones and preserve only those points that provide useful information. The higher the value displayed, the better that metric performs for rejection.

\section{Experiments and results}

This section describes the experiments carried out to contextualize the proposal and show the effectiveness of the proposed uncertainty wrapper.

\subsection{A sentiment analysis use case}
In order to illustrate the application of the method to a concrete problem, we propose a scenario in which a sentiment analysis system is applied to product reviews. The goal of the system is to classify each review on whether it is positive or negative. Later on, this classification can be used to, for example, use the positive reviews as implicit feedback for a recommender system.

The goal of the experiment is two-fold. First, we want to show how to apply the wrapper for a given NLP task. Second, we demonstrate how the proposed method additionally captures the uncertainty caused by the change in domains. To this end, we include different combinations of training and prediction domains in the experiment.
\subsection{Datasets}
The details on the datasets used are the following:
\begin{itemize}
    \item Stanford Sentiment Treebank \cite{socher-etal-2013-recursive}, SST-2, binary version where the purpose of the tasks is to classify a movie review in two categories: a positive or negative review. The dataset is split in 65,538 test samples, 872 for validation and 1,821 for testing.
    \item Yelp challenge 2013\footnote{\url{https://www.yelp.com/dataset/challenge}}, the goal is to classify reviews about Yelp venues where their users rated them using 1 to 5 stars. To be able to reuse a classifier trained with the SST-2 problem, we transform the Yelp dataset from a multiclass set to a binary one by grouping the ratings below three as a negative review, and positive otherwise. The dataset is split in 186,189 test samples, 20,691 for validation and 22,991 for testing.
    \item Amazon  Multi-Domain Sentiment dataset contains product reviews taken from Amazon.com from many product types (domains) \cite{blitzer-etal-2007-biographies}. As in Yelp, the dataset consists on ratings from 1 to 5 stars that we label as positive for those with values greater or equal to 3, and negative otherwise, split into train, validation and test datasets. We use two of the domains available: music (109,733/12,193/52,254 examples) and electronics (14,495/1,611/6,903 examples).  
\end{itemize}
\subsection{Experiment Set up}
On every experiment, we will be using two datasets: (i) a source dataset is used for training a model, that will be considered a black-box model from that moment on and (ii) a target dataset that corresponds to the domain we want to apply the black-box model on. It is in this second dataset that we will use the wrapper to measure the uncertainty. Specifically, the steps followed on each case are:
\begin{itemize}
	\item Train the black-box. First, we train a binary sentiment analysis classifier with the source dataset. In real scenarios, this step would not be necessary as we would be using a pre-trained model or third-party API. Because we train this model, we have more control over the experiment.
	\item Apply the black-box to the target domain. In this step, we use the black-box to obtain the predictions. In this stage, we can evaluate the accuracy of the target dataset, and we can compute the predictive entropy based on the prediction outputs.
	\item Compute the uncertainty for the target domain using the wrapper model. Once we have the predictions for the target domain, we proceed to train the uncertainty wrapper to approximate the Dirichlet probability density function for each input. By sampling the pdf, we compute the sampling predictive entropy of the average of the outputs to get the uncertainty score for each review in the target dataset.
	\item Apply the rejection mechanism. Finally, we use the uncertainty score to sort the predictions from more to less uncertain, and we search for a rejection point that maximizes the three performance measures: non-rejected accuracy, and classification and rejection quality.
\end{itemize}

\begin{figure}[ht!]
\centering
\caption{Model used to train the black-box models}
\includegraphics[scale=0.4]{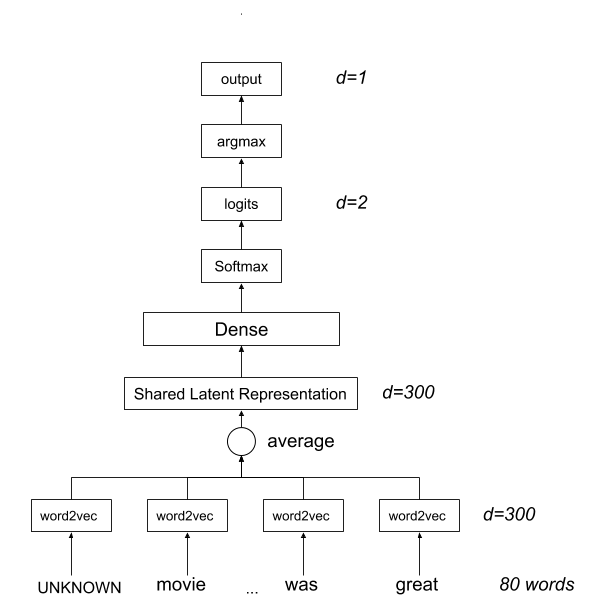}
\label{fig:bb_model}
\centering
\end{figure}

Figure \ref{fig:bb_model} describes the model used for training the black box models. As stated before, the only purpose of this model is to obtain a black-box classifier for a given source domain. The goal, in this case, is not to obtain the best classifier but to obtain a model which is easy to train and offers good performance.  We also want to remark that in this model we make use of a pre-trained set of word2vec embeddings, which is interesting to show how the model can be applied when there is a concatenation of different pre-trained models, each of them introducing their variability to the final output.

We run four different scenarios, namely:
\begin{itemize}
    \item {\bf Scenario 1:} The source dataset used for training the black-box is the {\it Yelp} dataset. The target domain for our application is given by the {\it SST-2} dataset.
    \item {\bf Scenario 2:} The source dataset used for training the black-box is the {\it SST-2} dataset. The target domain for our application is given by the {\it Yelp} dataset.
    \item {\bf Scenario 3:} The source dataset used for training the black-box is the {\it Amazon electronic products reviews} dataset. The target domain for our application is given by the {\it Amazon music products reviews} dataset.
    \item {\bf Scenario 4:} The source dataset used for training the black-box is the {\it Amazon music products reviews} dataset. The target domain for our application is given by the {\it Amazon electronic products reviews} dataset.
\end{itemize}

Each model is trained for 80 epochs, using Adam optimizer with a learning rate of $10^{-3}$.

\subsection{Results}

In order to show the effect of the application of the uncertainty wrapper on each of the target domains, we compute the uncertainty score using the three different metrics described in section 4: the predictive entropy of the black-box output ({\it baseline}), the predictive entropy obtained after training the aleatoric wrapper ({\it pred. entropy}), and the variation ratios ({\it var. ratios}). The first metric is taken by directly observing the output of the black-box (the entropy of the probabilities), while for obtaining the second and third metrics we train the aleatoric wrapper with optimal values for the learning rate, the number of epochs and batch size.

\begin{figure*}[!ht]
\centering
\minipage{1.0\textwidth}
  \centering
  \includegraphics[width=0.85\linewidth]{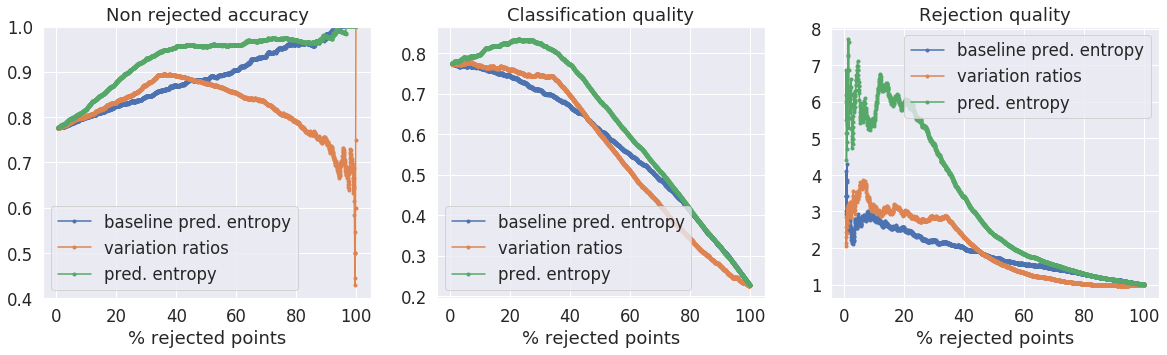}
  \caption{Apply Yelp BB to SST-2}\label{fig:yelptosst2}
\endminipage\hfill
\minipage{1.0\textwidth}
  \centering
  \includegraphics[width=0.85\linewidth]{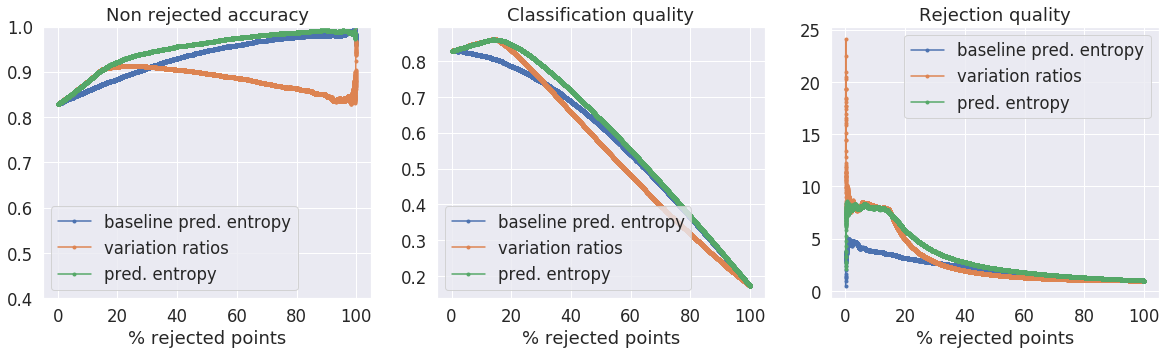}
  \caption{Apply SST-2 BB to Yelp}\label{fig:sst2toyelp}
\endminipage\hfill
\minipage{1.0\textwidth}
  \centering
  \includegraphics[width=0.85\linewidth]{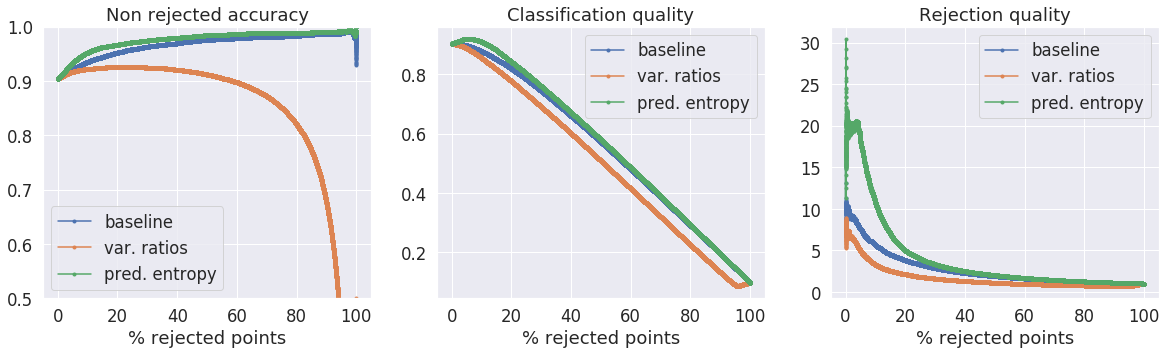}
  \caption{Apply electronics BB to Music}\label{fig:electrtomusic}
\endminipage\hfill
\minipage{1.0\textwidth}
  \centering
  \includegraphics[width=0.85\linewidth]{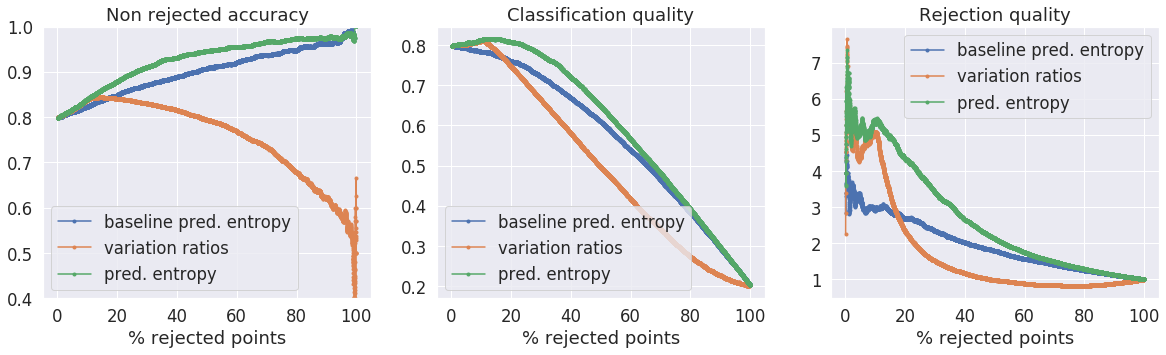}
  \caption{Apply music BB to electronics}\label{fig:musictoelectr}
\endminipage\hfill
\end{figure*}

Figures \ref{fig:yelptosst2} to \ref{fig:musictoelectr} show the results obtained on each combination for the rejection performance metrics. We plot  the three performance metrics of the rejection method proposed using each uncertainty score. The x-axis corresponds to the percentage of points rejected from the test set, always following the order of the uncertainty score, from more to less uncertain. In the y-axis, we plot the performance metric of the prediction of the preserved points. On the left-hand side, we observe the non-rejected accuracy plot. Classification quality is found at the centre. And, rejection quality on the right hand side. The higher the value in the plot, the better the result.

Observe that the proposed method displays better behaviour in all scenarios and metrics. This shows that the predictive entropy obtained through the aleatoric wrapper better captures the error of the predictions than by just using the entropy of the original output. Even though the proposed rejector outperforms the baseline in terms of the accuracy of the preserved data, one can observe that it depends on each scenario. We note that in those combinations where the adaptation of the model to the target domains is worst, the aleatoric predictive entropy is much better. We also observe that variation ratios fail in all metrics at capturing the error in this setup.

Analysing the classification quality (plot at the center of each figure), we can see that the proposed metric is also excellent at separating right from wrong points, rejecting the misclassified ones. Finally, observing the rejection quality, we can see the ability of the uncertainty score in rejecting misclassified samples. Again, the proposed method clearly displays better behavior than the baseline and variation ratios.

Table \ref{tab:results} shows the numerical results obtained during the experiments for the four combinations tested. The first column, black-box source acc, describes the accuracy obtained for the source dataset after training the original classifier. Next, column black-box target acc describes the accuracy obtained when applying the black-box to the target dataset. The rest of the columns show the non-rejected accuracy and the classification and rejection quality after rejecting 10, 20 an 30\% of the points, using the proposed predictive entropy as a rejector.

\begin{table*}[ht!]
\large
\centering
\caption{Accuracy obtained by training an standalone classifier, applying the API and the proposed wrapper for each domain}
\label{tab:results}
\resizebox{\linewidth}{!}{%
\begin{tabular}{|l|c|c|c|c|c|}
\hline
 & \textbf{\begin{tabular}[c]{@{}c@{}}BB source \\ acc.\end{tabular}} & \textbf{\begin{tabular}[c]{@{}c@{}}BB target \\ acc.\end{tabular}} & \textbf{\begin{tabular}[c]{@{}c@{}}Non-reject. \\ acc.\\ (10/20/30\%)\end{tabular}} & \textbf{\begin{tabular}[c]{@{}c@{}}Class. \\ quality\\ (10/20/30\%)\end{tabular}} & \textbf{\begin{tabular}[c]{@{}c@{}}Reject. \\ quality\\ (10/20/30\%)\end{tabular}} \\ \hline
\textbf{Apply Yelp BB to SST-2} & 89.18$\pm$0.08\% & 77.13$\pm$0.52\% & \begin{tabular}[c]{@{}c@{}}82.43$\pm$0.22\%\\ 88.19$\pm$0.50\%\\ 93.60$\pm$0.16\%\end{tabular} & \begin{tabular}[c]{@{}c@{}}80.40$\pm$0.39\%\\ 83.11$\pm$0.80\%\\ 83.05$\pm$0.23\%\end{tabular} & \begin{tabular}[c]{@{}c@{}}6.03$\pm$0.45\\ 6.04$\pm$0.51\\ 4.97$\pm$0.07\end{tabular} \\ \hline
\textbf{Apply SST-2 BB to Yelp} & 83.306$\pm$0.18\% & 82.106$\pm$0.88\% & \begin{tabular}[c]{@{}c@{}}87,98$\pm$0.18\%\\ 92.13$\pm$0.38\%\\ 94.19$\pm$0.33\%\end{tabular} & \begin{tabular}[c]{@{}c@{}}85.49$\pm$0.88\%\\ 84.53$\pm$0.38\%\\ 78.99$\pm$0.46\%\end{tabular} & \begin{tabular}[c]{@{}c@{}}8.30$\pm$1.63\\ 5.72$\pm$0.27\\ 3.73$\pm$0.10\end{tabular} \\ \hline
\textbf{Apply Electronics BB to Music} & 86.39$\pm$0.22\% & 90.38$\pm$0.13\% & \begin{tabular}[c]{@{}c@{}}95.04$\pm$0.43\%\\ 96.45$\pm$0.35\%\\ 97.26$\pm$0.31\%\end{tabular} & \begin{tabular}[c]{@{}c@{}}90.67$\pm$0.88\%\\ 83.93$\pm$0.67\%\\ 75.77$\pm$0.54\%\end{tabular} & \begin{tabular}[c]{@{}c@{}}10.7$\pm$1.65\\ 4.82$\pm$0.35\\ 3.25$\pm$0.14\end{tabular} \\ \hline
\textbf{Apply Music BB to Electronics} & 93.10$\pm$0.02\% & 79.85$\pm$0.0\% & \begin{tabular}[c]{@{}c@{}}83.26$\pm$0.41\%\\ 87.06$\pm$0.55\%\\ 90.50$\pm$0.29\%\end{tabular} & \begin{tabular}[c]{@{}c@{}}79.97$\pm$0.74\%\\ 79.93$\pm$0.87\%\\ 76.81$\pm$0.41\%\end{tabular} & \begin{tabular}[c]{@{}c@{}}4.1$\pm$0.55\\ 3.80$\pm$0.35\\3.32$\pm$0.09\end{tabular} \\ \hline
\end{tabular}%
}
\end{table*}

In general terms, the results displayed in table \ref{tab:results} show that the rejection method can reduce the error of the output predictions when applying a pre-trained black-box classification system to a new domain. It is remarkable that in all the use cases, by rejecting only a 10\% of the more uncertain predictions, the resulting system increases its accuracy between four and eight percentual points.

Results demonstrate how the usage of the uncertainty for rejecting uncertain predictions helps with the adaptation of a pre-trained model to new domains of application. In some cases, the results obtained for the test dataset of the target domain by rejecting 10\% of the less certain points overtake those obtained by the source dataset used for training the original model. As a curiosity, the use case where we trained a black-box model using the reviews of Amazon's electronics products achieves better results when applied to the test target dataset than to the original test dataset. Even in this case, where the applied classifier reaches an accuracy of more than 90 \%, the proposed method increases it in almost 5 points. 

The percentage of rejected samples will vary on each use case, depending on the feasibility of rejecting prediction points, but it is clear that the proposed method improves the performance of the resulting system in terms of the accuracy of the predictions leading to more reliable and confident decisions based on them. 

With regards to the interpretability of the uncertainty scores obtained, we performed a qualitative analysis of the results obtained. The analysis of the more uncertain predictions discarded point to potential causes of high uncertainty as the inclusion of negative words, like "awful" and "disappointed" in very positive reviews (and the other way around), or ambiguous reviews that hold positive and negative tenses at the same time.

These results illustrate a practical case in which a wrapper technique is built to provide uncertainty scores in the predictions of a simulated black-box model. As a result, the black-box model is more accountable with respect to the accuracy dimension and actions can be taken for risk mitigation in cases where the decision is not certain enough. Observe that by using rejection as a risk mitigation mechanism, we obtain a more trustable system with higher accuracy performance.

\section{Conclusions and Future Work}

Trustable machine learning tools require algorithm accountability and transparency. Despite this need, most of the current data-based solutions fail to certify those features. One of the reasons for that is their architecture complexity, as these products usually depend on third-party black-box components or SaaS APIs, among others.

In this work, we introduce a wrapper technique that is able to take any black-box solution and endow it with uncertainty features. By using this wrapper, we make the black-box auditable for the accuracy risk (risk derived from low quality or uncertain decisions) while also providing an actionable mechanism to mitigate that risk in the form of decision rejection (we may opt not to output a prediction when the risk or uncertainty in that decision is significant).

To that end, we propose a wrapper based on the Dirichlet distribution that is able to capture aleatoric uncertainty when a black-box solution is applied to our problem domain. We quantify that uncertainty and leverage this to develop a rejection system to increase the accuracy and confidence in the resulting system. By rejecting the more uncertain predictions, the predictor reduces the potential adverse effects of making decisions based on groundless evidence.

To showcase the benefits of the method and describe how to implement it in a particular problem, we apply it to a sentiment analysis problem on a Natural Language Processing scenario. We use an immutable simulated API to obtain the sentiment for a corpus of products reviews that belong to a domain different from that used to train the API. Results show how the method effectively selects uncertain predictions, and by omitting these, we improve the accuracy of the API for our problem in 4 to 8 points by only rejecting up to 10\% of the samples. More significant improvements are achieved as the number of rejected points increases.

Even though the experiments included in this work are limited to binary classifiers, the proposed uncertainty wrapper is not constrained in the number of classes. In the next steps, we plan to apply the method to more complex NLP problems that include multiclass predictions, like language modelling or question answering.  

Additionally, we plan to explore different architectures for the implementation of the wrapper and see how the different models can help on better capturing the uncertainty. We also plan to analyse the role of other types of uncertainty, like the epistemic or model uncertainty derived from the use of other models, and to approximate the uncertainty derived by out-of-sampling data points.

\paragraph{Acknowledgements}
This work has been partially funded by the Spanish projects TIN2016-74946-P, TIN2015-66951-C2 (MINECO/FEDER, UE), RTI2018-095232-B-C21 and 2017 SGR 1742, and by AGAUR of the Generalitat de Catalunya through the Industrial PhD grant.

\bibliographystyle{unsrt}  
\bibliography{dirichlet_uncert_wrappers}

\end{document}